\documentclass[conference]{IEEEtran}
\IEEEoverridecommandlockouts  

\usepackage[activate={true,nocompatibility}]{microtype}
\usepackage{graphicx}
\usepackage[font=footnotesize]{caption}
\usepackage{titlesec}
\usepackage{subcaption}
\usepackage{booktabs}
\usepackage{amsmath}
\usepackage{amssymb}
\usepackage{algorithm}
\usepackage{algorithmic}
\usepackage{hyperref}
\hypersetup{hidelinks}
\usepackage{tikz}
\usetikzlibrary{arrows.meta, positioning, calc, shapes.geometric, fit, backgrounds}
\usepackage{xcolor}
\usepackage{placeins}  
\usepackage{float}     
\usepackage{stfloats}  

\graphicspath{{images/}{flowcharts/}{Images/}}

\tikzset{
  vaeblock/.style={rectangle, rounded corners=3pt, draw=black!55,
    line width=0.55pt, align=center, font=\footnotesize,
    minimum height=11mm, minimum width=16mm, inner xsep=5pt, inner ysep=4pt},
  vaeencoder/.style={vaeblock, fill=blue!8,   draw=blue!55},
  vaepooler/.style ={vaeblock, fill=teal!10,  draw=teal!60},
  vaelatent/.style ={vaeblock, fill=violet!10,draw=violet!60},
  vaedecoder/.style={vaeblock, fill=orange!12,draw=orange!70},
  vaeio/.style     ={vaeblock, fill=gray!8,   draw=gray!60,
                     minimum height=11mm, minimum width=14mm},
  vaearrow/.style  ={-{Stealth[length=4.5pt, width=3.5pt]}, line width=0.6pt, draw=black!60},
  vaetensor/.style ={font=\scriptsize, text=black!65}
}

\title{CloudDiffusion: Diffusion-Based Scene Completion\\in the Point Cloud Domain}

\author{%
\IEEEauthorblockN{%
Chidera Agbasiere\textsuperscript{*},\,
Mikhail Sannikov\textsuperscript{*},\,
Faith Ogunwoye,\,
Erik Shaikhiev,\\
Alex Kozinov,\,
Ilya Mikhalchuk,\,
Iana Zhura,\,
and Dzmitry Tsetserukou%
}
\thanks{\textsuperscript{*}These authors contributed equally to this work.}
\thanks{All authors are with the Intelligent Space Robotics Laboratory, Skolkovo Institute of Science and Technology, Moscow, Russia. \{chidera.agbasiere, mikhail.sannikov, faith.ogunwoye, erik.shaikhiev, alex.kozinov, ilya.mikhalchuk, iana.zhura, d.tsetserukou\}@skoltech.ru}
}

\begin{document}

\maketitle


\begin{abstract}
Reconstructing dense 3D scenes from sparse LiDAR point clouds (LiDAR scene
completion) is a fundamental
challenge in autonomous driving, where diffusion models offer a promising
solution. However, existing approaches rely on object-level autoencoders that
collapse into unstable global representations at outdoor scale, and suffer from
ground truth data corrupted by odometry drift that systematically degrades
supervision quality. Furthermore, multi-step diffusion inference incurs
prohibitive latency for real-time deployment.
We present \emph{CloudDiffusion}, addressing these issues with three
independent components. First, a multi-token
Gaussian VAE with cross-attention pooling provides stable scene-scale LiDAR
compression as a \emph{standalone} reconstruction module, avoiding the
global-pooling and codebook-collapse failure modes of prior point-cloud
autoencoders. Second, an anchor-based ICP ground truth refinement pipeline
eliminates drift-induced noise from training supervision, reducing our
single-step $x_0$ diffusion teacher's squared Chamfer distance by
${\sim}16\times$ on SemanticKITTI seq.~08 ($0.396 \to 0.024$\,m$^2$) with no
model change, largely reflecting the denser, more compact refined
references (${\approx}8\%$ gain at a fixed reference).
Third, the same teacher completes scenes in a single $x_0$ step from a
noised target-derived coordinate scaffold, operating
\emph{directly in coordinate space}, not in the VAE latent. It runs in near
real time at $209$\,ms/frame, $23$--$98\times$ lower inference latency than
iterative diffusion baselines. Our results indicate that data quality dominates model design in
this regime, and suggest that multi-token latent spaces could serve as a
stable first stage for future latent diffusion-based scene completion.
Code, checkpoints, and the ground-truth pipeline:
\href{https://github.com/A-C-Simon/sonata_ws}{\texttt{github.com/A-C-Simon/sonata\_ws}}.
\end{abstract}

\section{Introduction}

\begin{figure}[!t]
\centering
\includegraphics[width=\columnwidth]{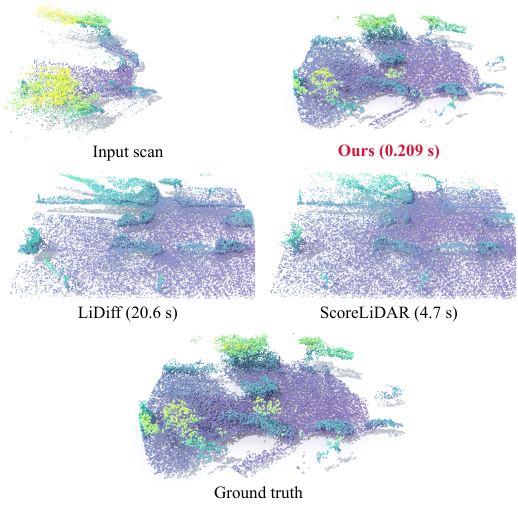}
\caption{\textbf{Single-step LiDAR scene completion vs.\ iterative
diffusion baselines.} From a partial LiDAR scan (top left), our $x_0$
diffusion teacher completes the scene in a single $209$\,ms forward pass
from a noised target-derived coordinate scaffold (Sec.~\ref{sec:teacher};
top right), closely matching the refined ground truth (bottom). LiDiff
($20.6$\,s) and ScoreLiDAR ($4.7$\,s), run on the same RTX~4090 without
such a scaffold, are $23$--$98\times$ slower; both complete the full
$360^\circ$ scene toward the original aggregated ground truth (shown
cropped to the evaluated volume), hence their fuller footprint and
smoother geometry.
All clouds rendered at an equal $20$k-point budget, coloured by height;
SemanticKITTI seq.~08, frame~3000.}
\label{fig:hero}
\end{figure}

Autonomous navigation demands dense 3D scene understanding from inherently
sparse sensor inputs.
A single LiDAR scan captures high-precision geometry but leaves large
fractions of the scene unobserved due to occlusion, range limits, and
angular sampling gaps.
Completing these missing regions is essential for downstream planning and
collision avoidance; our method performs this completion in a single forward
pass (Fig.~\ref{fig:hero}).
\begin{figure*}[t]
\centering
\includegraphics[width=0.66\textwidth]{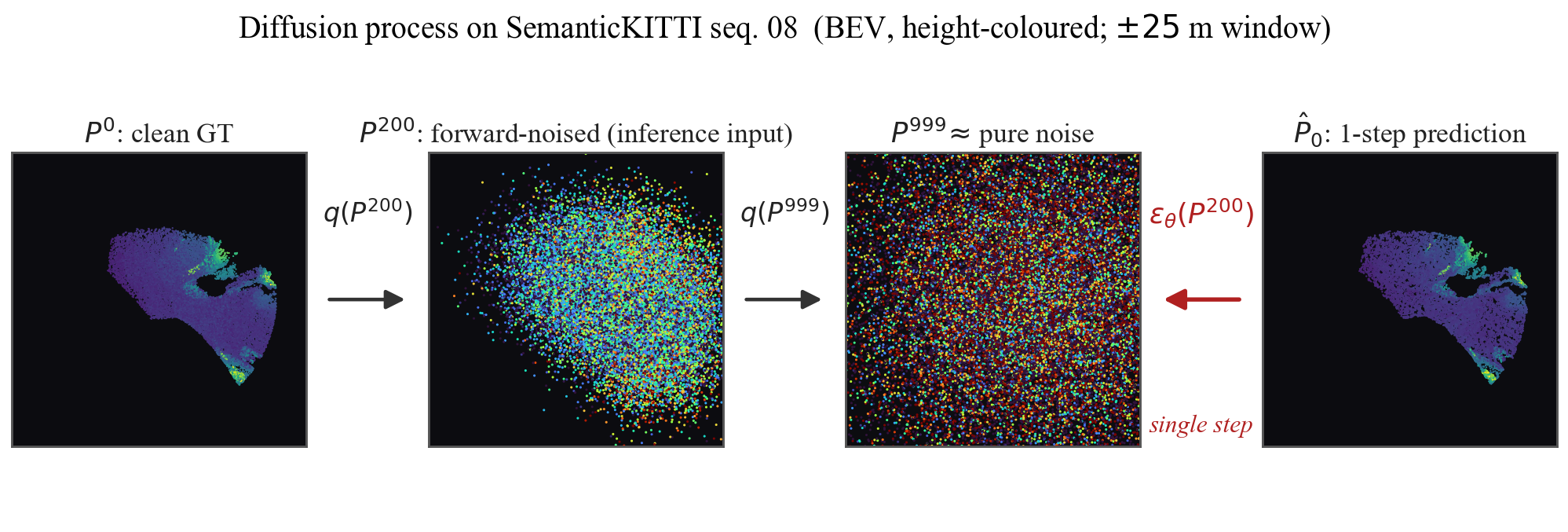}
\caption{Diffusion process on a SemanticKITTI seq.\ 08 scene
(BEV, height-coloured, $\pm 25$\,m window). \textbf{Left to right:} clean
target $P^0$; forward-noised $P^{200}$ (our single-step inference input);
near-pure noise $P^{999}$; prediction $\hat{P}_0$ recovered from $P^{200}$
in one forward pass (Sec.~\ref{sec:teacher}).}
\label{fig:diffusion_process}
\end{figure*}

Diffusion-based generative models are well suited to this task: they model
a distribution over plausible completions rather than a single deterministic
estimate, a natural fit for the one-to-many ambiguity of occluded
geometry.
Beyond completion, diffusion models have rapidly become a versatile generative
prior across robotics, spanning navigation~\cite{EmbodiedDiffusion2026,DiffusionAnything2026},
manipulation and grasping~\cite{DiffusionRL2026}, and aerial path
planning~\cite{ImpedanceDiffusion2026}; this motivates their use for dense 3D
perception.

This power comes at a cost: tens to hundreds of denoising steps push
per-frame latency far outside the real-time budget of a moving vehicle,
which must refresh its surroundings within hundreds of milliseconds.
Collapsing inference to a single step while retaining fidelity is
therefore a primary design goal of this work.

Outdoor scale also motivates a compact latent code that compresses tens
of thousands of unordered points; our standalone VAE targets this role as
a candidate first stage for future latent-diffusion completion.

Existing point-cloud autoencoders offer either a single global pooling
vector or discrete codebooks prone to collapse at scene scale; both fail
as latent-diffusion foundations.

We make three primary contributions:
\begin{enumerate}
  \item \textbf{Multi-token Gaussian VAE.}
    A PointNet encoder feeds a cross-attention pooler that extracts
    32~spatially specialised latent tokens.
    A transformer decoder with 5 cross-attention blocks reconstructs
    8{,}000 scene points, achieving squared Chamfer distance
    $0.120\pm0.026$\,m$^2$ (50 validation frames) at 7.1\,M parameters and
    1.6\,ms inference per frame.
  \item \textbf{Drift-reduced ground truth construction.}
    Anchor-based ICP alignment of aggregated scans reduces the diffusion
    teacher's squared Chamfer distance from $0.396 \pm 0.090$\,m$^2$ to
    $0.024 \pm 0.005$\,m$^2$ (a ${\sim}16\times$ gain) with no change to
    the model; most of the ratio reflects the denser, tighter references
    (Sec.~\ref{sec:discussion}).
  \item \textbf{Single-step coordinate-space completion.}
    Our diffusion teacher completes a scene in a single $x_0$ forward pass
    directly in coordinate space, at $209$\,ms/frame on an RTX~4090.
    This is $23\text{--}98\times$ lower inference latency than iterative
    diffusion baselines (LiDiff, ScoreLiDAR), making diffusion-based LiDAR
    completion practical for near-real-time deployment.
\end{enumerate}

\section{Related Work}

\textbf{Scene Completion.}
Scene completion recovers dense geometry from a single sparse LiDAR scan.
Early voxel-grid methods (LMSCNet~\cite{LMSCNet2020}) regress occupancy
deterministically, fast but resolution-capped. Camera-conditioned semantic scene completion
(VoxFormer~\cite{VoxFormer2023}, CGFormer~\cite{CGFormer2024},
SGN~\cite{SGN2024}, DepthSSC~\cite{DepthSSC2025}) lifts image features into 3D
but remains bounded by the underlying depth estimator. Implicit and self-supervised approaches
(LODE~\cite{LODE2023}) relax the voxel constraint yet still
commit to a single deterministic surface, and the recent non-diffusion
LiNeXt~\cite{LiNeXt2025} leads in single-pass efficiency. The shared limitation is
determinism: a single committed completion blurs ambiguous occluded
regions.

\textbf{Diffusion-Based 3D Scene Completion.}
Diffusion models instead sample a distribution over plausible
completions. R2DM~\cite{R2DM2024} and LiDMs~\cite{LiDMs2024}
generate high-fidelity scans from range images; DiffComplete~\cite{DiffComplete2024}
adapts diffusion to shape completion, while LiDiff~\cite{LiDiff2024} operates
directly on points and redefines the noise schedule to set the current quality
frontier. To attack the dominant weakness, latency,
ScoreLiDAR~\cite{ScoreLiDAR2025} distills LiDiff into a few-step student,
cutting per-frame time from ${\sim}30$\,s to ${\sim}5$\,s while improving
Chamfer distance; Distillation-DPO~\cite{DistDPO2025} pushes distillation
further with preference optimisation, though completion remains multi-second. LiDPM~\cite{LiDPM2025} iterates point-space sampling from the scan noised
to an intermediate timestep; we instead make one $x_0$ prediction at a fixed
timestep. Even distilled models remain
multi-step (typically 8) and bounded by their teacher. Our single-step $x_0$
teacher targets exactly this gap at $209$\,ms/frame.

\textbf{Point Cloud Autoencoders.}
Global max-pooling architectures (PointNet~\cite{PointNet2017}) produce a single
latent vector that discards fine spatial structure.
VQ-VAEs~\cite{VQVAE2017} introduce discrete codebooks but suffer from collapse when the
scene distribution is broad and multi-peaked, as in outdoor LiDAR.
Perceiver-style vec-set latents~\cite{Perceiver2021,Shape2VecSet2023}
have been used for object-scale 3D generation, but their behaviour at
outdoor driving scale is under-studied.
Our multi-token Gaussian VAE transfers this design to scene-scale LiDAR
and avoids both failure modes by maintaining 32 spatially specialised,
continuous Gaussian tokens.

\textbf{Ground Truth Quality.}
Prior work assumes accumulated LiDAR maps are reliable supervision sources.
nuCraft~\cite{nuCraft2024} plays this role for nuScenes occupancy;
ours is the SemanticKITTI-completion analogue.
We show that pose drift in standard SemanticKITTI poses inflates Chamfer
distance substantially, and that local ICP correction is sufficient
for stable diffusion training.

\section{Method}

\subsection{Overview}

Our scene-completion pipeline operates entirely in coordinate (point) space
and has two stages:
\begin{enumerate}
  \item A frozen Point Transformer V3 (PTv3/Sonata,
    108\,M parameters)~\cite{PTv3,Sonata} encodes the partial LiDAR scan into
    256-dim conditioning features.
  \item A lightweight denoising network (8.9\,M parameters) is trained by
    diffusion \emph{directly in coordinate space}, conditioned on the PTv3
    features, and outputs the completed point cloud; at inference it uses a
    single $x_0$ prediction step.
\end{enumerate}

Separately, we develop a multi-token Gaussian VAE (7.1\,M parameters) as a
\emph{standalone} reconstruction module that compresses a dense scene into
32~continuous latent tokens and decodes it back to points. The VAE and the
diffusion teacher are trained \emph{independently} and serve distinct
roles: the teacher performs sparse-to-dense completion in coordinate space,
while the VAE is studied as a stable scene-scale latent representation, a
candidate first stage for future \emph{latent}-diffusion completion. In the
present system the VAE is \emph{not} used as a bottleneck inside the diffusion
teacher; integrating the two is left to future work.

\begin{figure*}[t]
\centering
\resizebox{1\linewidth}{!}{%
\begin{tikzpicture}[node distance=10mm and 9mm]
  \node[vaeio]                                       (in)    {Input\\[-1pt]{\scriptsize $P\!\in\!\mathbb{R}^{N\times 3}$}};
  \node[vaeencoder, right=of in]                     (mlp)   {PointNet MLP\\[-1pt]{\scriptsize $f_i\!\in\!\mathbb{R}^{256}$}};
  \node[vaepooler,  right=of mlp]                    (cross) {Cross-Attn\\Pooler\\[-1pt]{\scriptsize 32 queries}};
  \node[vaelatent,  right=22mm of cross]              (mu)    {Posterior\\$(\mu_k,\log\sigma_k^2)$\\[-1pt]{\scriptsize $k{=}1{:}32$}};
  \node[vaelatent,  right=of mu]                     (z)     {Reparam.\\$z_k{=}\mu_k{+}\sigma_k\!\odot\!\epsilon_k$};

  \draw[vaearrow] (in)    -- (mlp);
  \draw[vaearrow] (mlp)   -- (cross);
  \draw[vaearrow] (cross) -- node[vaetensor, above, yshift=1pt]{$T\!\in\!\mathbb{R}^{32\times 256}$} (mu);
  \draw[vaearrow] (mu)    -- (z);

  \coordinate (row1mid) at ($(mlp.south)!0.5!(z.south)$);
  \node[vaedecoder, anchor=north] (dec) at ([yshift=-14mm, xshift=12mm]row1mid)
        {Decoder\\$5\!\times$[CrossAttn\,+\,MLP]\\[-1pt]{\scriptsize 8{,}000 queries}};
  \node[vaeio, anchor=east] (out) at ([xshift=-22mm]dec.west)
        {Output\\[-1pt]{\scriptsize $\hat{X}\!\in\!\mathbb{R}^{8000\times 3}$}};

  \draw[vaearrow] (z.south)  -- ++(0,-6mm) -| (dec.north);
  \draw[vaearrow] (dec.west) -- node[vaetensor, above, yshift=1pt]{$\hat{X}$} (out.east);

  \begin{scope}[on background layer]
    \node[draw=blue!35, dashed, rounded corners=4pt, inner xsep=3mm, inner ysep=4mm,
          fit=(mlp)(cross),
          label={[font=\scriptsize\itshape\color{blue!60}, yshift=-2pt]above:Encoder}] {};
    \node[draw=violet!45, dashed, rounded corners=4pt, inner xsep=3mm, inner ysep=4mm,
          fit=(mu)(z),
          label={[font=\scriptsize\itshape\color{violet!60}, yshift=-2pt]above:{Gaussian Latent (32 tokens $\times$ 32-d)}}] {};
    \node[draw=orange!55, dashed, rounded corners=4pt, inner xsep=3mm, inner ysep=4mm,
          fit=(dec),
          label={[font=\scriptsize\itshape\color{orange!70}, yshift=2pt]below:Decoder}] {};
  \end{scope}
\end{tikzpicture}%
}

\caption{Multi-token Gaussian VAE: a PointNet MLP and a
32-query cross-attention pooler produce per-token Gaussian posteriors
($32\times32$-d latent); 8{,}000 learned queries cross-attend through 5
decoder blocks to reconstruct $\hat{X}$ (Sec.~\ref{sec:vae}).}
\label{fig:vae_arch}
\end{figure*}
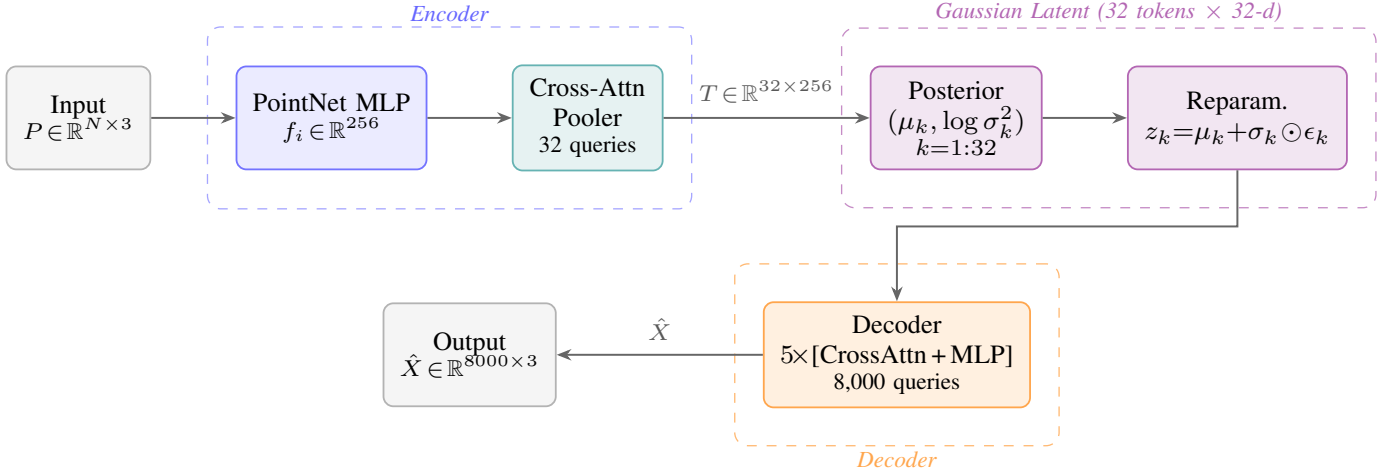
\subsection{Multi-Token Gaussian VAE}
\label{sec:vae}

\subsubsection{Encoder}

The full architecture is shown in Fig.~\ref{fig:vae_arch}.
A PointNet-style MLP~\cite{PointNet2017} processes each input point independently:
\begin{equation}
  f_i = \mathrm{MLP}_3(\mathrm{MLP}_2(\mathrm{MLP}_1(xyz_i))),
  \quad f_i \in \mathbb{R}^{256}.
\end{equation}
A cross-attention pooler with 32 learned query vectors extracts
spatially specialised global tokens:
\begin{equation}
  T = \mathrm{CrossAttn}(\mathrm{Queries},\, F),
  \quad T \in \mathbb{R}^{32 \times 256},
\end{equation}
where $F=[f_1;\dots;f_N]$ stacks the per-point features. Each token $T_k$ is projected to a Gaussian posterior with
$\mu_k, \log\sigma^2_k \in \mathbb{R}^{32}$ (so that
$\sigma_k = \exp(\tfrac{1}{2}\log\sigma_k^2)$).

\subsubsection{Latent Representation}

The 32 tokens of 32 dimensions yield a compact $1{,}024$-dimensional
latent. Each latent token $z_k$ is sampled from its posterior with the
reparameterisation trick,
\begin{equation}
  z_k = \mu_k + \sigma_k \odot \epsilon_k,
  \quad \epsilon_k \sim \mathcal{N}(0, I),
\end{equation}
where $\mu_k$ and $\sigma_k$ are the posterior mean and standard
deviation of token $k$, $\epsilon_k$ is standard Gaussian noise, and
$\odot$ denotes element-wise multiplication.

\subsubsection{Decoder}

The decoder's 8{,}000 learned point queries cross-attend to the latent tokens through
5 transformer decoder blocks, each consisting of cross-attention
and an MLP:
\begin{equation}
  \hat{X} = \mathrm{DecoderBlocks}(\mathrm{Queries},\, Z),
  \quad \hat{X} \in \mathbb{R}^{8000 \times 3},
\end{equation}
where $Z=[z_1;\dots;z_{32}]$ stacks the sampled latent tokens.

\subsubsection{Training Objectives}

We train the VAE with a symmetric squared-L2 Chamfer reconstruction
loss $\mathcal{L}_{\mathrm{CD}}$ between decoded points $\hat{X}$ and
ground-truth points $X_{\mathrm{gt}}$, averaged over both directions
to penalise missing and spurious geometry symmetrically. A per-token
Kullback--Leibler divergence $\mathcal{L}_{\mathrm{KL}}$ regularises
each of the 32 latent tokens against $\mathcal{N}(0,I)$ to prevent
posterior collapse while the reconstruction term dominates:
\begin{equation}
  \mathcal{L}_{\mathrm{VAE}} = \mathcal{L}_{\mathrm{CD}}
  + \beta\, \mathcal{L}_{\mathrm{KL}},
\end{equation}
\begin{equation}
\begin{split}
\mathcal{L}_{\mathrm{CD}}
 = \tfrac{1}{2}\Bigl[
  &\tfrac{1}{|X_{\mathrm{gt}}|}\!\sum_{x\in X_{\mathrm{gt}}}\!
     \min_{\hat{x}\in\hat{X}}\|x-\hat{x}\|^2 \\
  +\ &\tfrac{1}{|\hat{X}|}\!\sum_{\hat{x}\in\hat{X}}\!
     \min_{x\in X_{\mathrm{gt}}}\|x-\hat{x}\|^2
  \Bigr],
\end{split}
\end{equation}
\begin{equation}
\mathcal{L}_{\mathrm{KL}} = -\tfrac{1}{2}\sum_{k=1}^{32}\sum_{d=1}^{32}\!
  \bigl(1+\log\sigma_{k,d}^2-\mu_{k,d}^2-\sigma_{k,d}^2\bigr),
\end{equation}
with $\beta=10^{-3}$. Before encoding, each point cloud is per-frame
mean-centred and isotropically rescaled to $[-1,1]$; the inverse
transform is applied after decoding so Chamfer distances are reported
in metres.

\subsection{Diffusion Teacher Pipeline}
\label{sec:teacher}

The PTv3 encoder processes the voxelised partial scan
(voxel size $0.05$\,m, up to 20{,}000 points) and produces
256-dim conditioning features.
A lightweight denoising network (8.9\,M parameters) is trained with
$\epsilon$-prediction~\cite{DDPM2020} under a cosine noise
schedule~\cite{iDDPM2021} (1{,}000 timesteps; AdamW, lr $10^{-4}$,
batch~2, fp16, 30 epochs).
Inference uses single-step $x_0$ prediction at $t{=}200$
($\bar\alpha_{200}\approx 0.897$; signal-to-noise ratio
$\mathrm{SNR}\approx 8.7$), which we found more
stable than full DDPM reverse sampling in our setting;
Fig.~\ref{fig:diffusion_process} illustrates the forward-noising process and
the single-step recovery. Note that inference predicts at coordinates
obtained by noising the completion target: reported completion quality is
conditioned on this coordinate scaffold, not recovered from the partial
scan alone.

\begin{figure*}[t]
\centering
\includegraphics[width=0.94\textwidth]{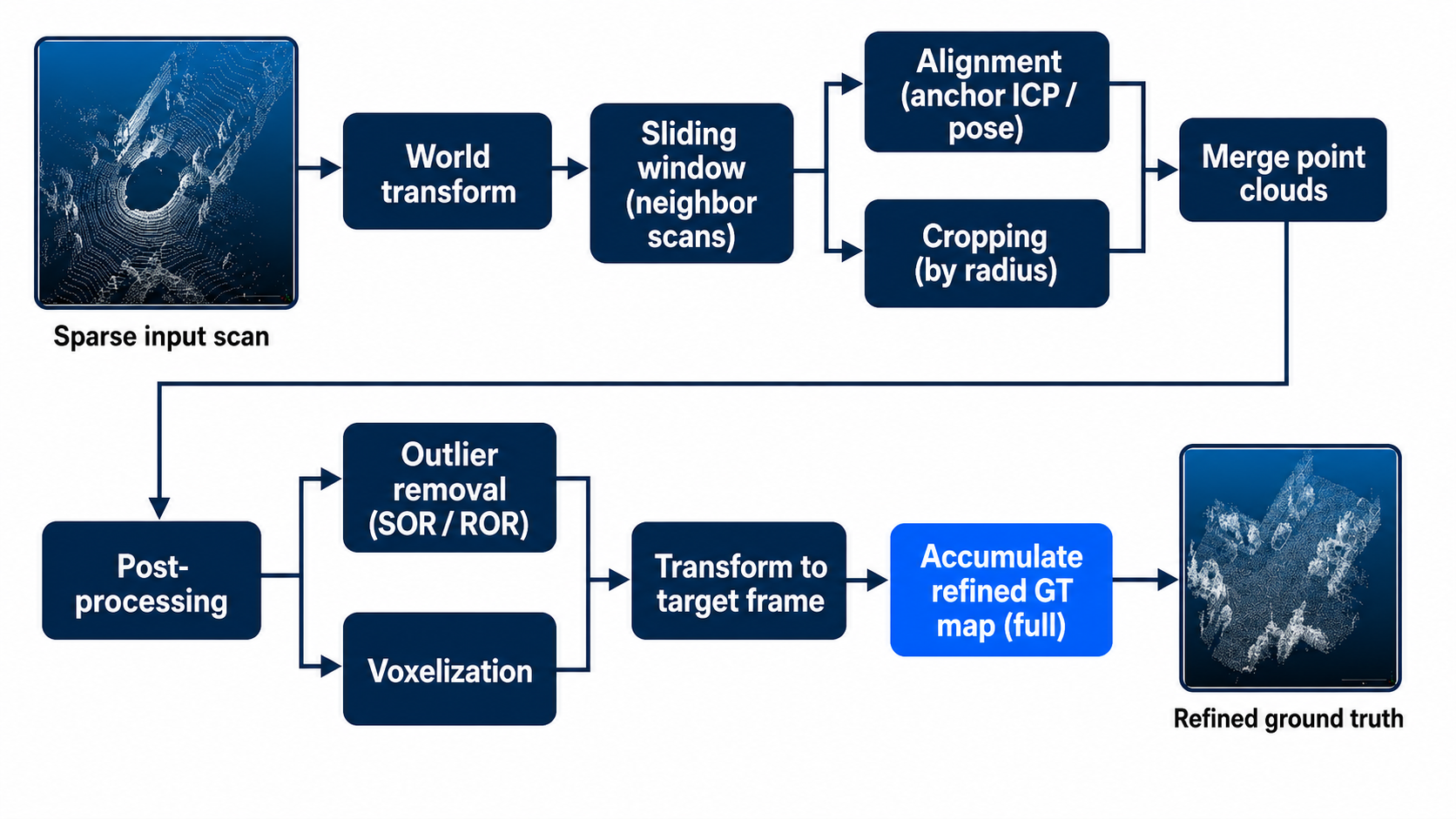}
\caption{Drift-reduced ground-truth construction: neighbouring scans are placed by pose, gathered over a sliding window, aligned by anchor ICP, cropped, and merged; SOR/ROR outlier removal, voxelisation, and transformation to the anchor frame yield the refined GT cloud.}
\label{fig:grtr_scheme}
\end{figure*}

\subsection{Drift-Reduced Ground Truth Construction}
\label{sec:gt}

Standard SemanticKITTI ground truth is built by accumulating sequential
scans using the dataset's SLAM-estimated poses.
Pose drift introduces systematic errors: doubled surfaces, thick walls,
and elevated Chamfer variance that destabilises diffusion training.

\textbf{Anchor-based ICP.}
For each anchor scan $s_i$ we aggregate a temporal window of neighbouring
scans $\{s_{i-w}, \ldots, s_{i+w}\}$ with $w=17$ (a 35-scan context), where each
scan $s_j \subset \mathbb{R}^{3}$ is a point set with $|s_j|=N_j$. Let
$G_j \in SE(3)$ be the dataset pose of scan $s_j$. Each scan is first cropped
to a $15$\,m radius about its own ego-position and placed in the shared world
frame by its pose, $s_j^{(0)} = G_j\, s_j$, and the window is processed in
temporal order: every newly placed scan is refined by point-to-point ICP
against a rolling reference $R_j$ formed from the $r{=}3$ most recently
processed scans (whether ICP-refined or pose-only; clouds voxel-downsampled to
$0.35$\,m, $\le 4$ iterations, $1.0$\,m maximum correspondence distance):
\begin{equation}
  \hat{G}_j = \mathrm{ICP}\!\left(s_j^{(0)},\, R_j\right).
\end{equation}
The ICP transform $\hat{G}_j \in SE(3)$ is rejected outright if its motion is
large (translation $>0.5$\,m or rotation $>5^\circ$), and otherwise accepted
only if the mean
per-point displacement $\Delta_j$ it induces stays below a threshold
$\tau = 0.15$\,m; when rejected, the pose-only placement $s_j^{(0)}$ is kept,
capping ICP to small drift corrections:
\begin{equation}
  \Delta_{j} = \frac{1}{|s_j^{(0)}|}\!\sum_{x \in s_j^{(0)}}
    \bigl\|\hat{G}_{j}\, x - x\bigr\| < \tau.
\end{equation}
The placed scans are accumulated, cropped to a $20$\,m radius about the anchor
ego-pose, voxelised at $0.1$\,m, and cleaned with statistical outlier removal
(SOR; $k{=}10$ neighbours, $2.0\sigma$) followed by radius outlier removal
(ROR; $\ge 5$ points within $0.5$\,m); the final map is expressed in the
anchor frame ($G_i^{-1}$). As the underlying scans are restricted to the
camera frustum, the accumulated reference spans a forward-facing volume.
Within its tighter extent the refined dataset is considerably denser than the
original accumulation cropped to the same volume, across $23{,}201$ frames
spanning the full SemanticKITTI training + validation split.

\section{Experiments}

\subsection{Setup}

\textbf{Dataset.}
We train on SemanticKITTI~\cite{SemanticKITTI2019} sequences 00--07 and
09--10, and validate on sequence~08. Throughout this paper a \emph{frame} denotes a single
LiDAR scan at one KITTI timestep together with its associated completion
target; all per-frame metrics are computed on that unit.
Diffusion teacher results (Tables~\ref{tbl:gt},~\ref{tbl:cd_compare}) are
evaluated on the full seq.~08 validation set (4{,}071 frames) with a
20{,}000-point subsample per frame.
The evaluation metric is the symmetric Chamfer distance (squared, in m$^2$, following
the PVD / LiDiff convention; lower is better). Linear CD (m) is reported
where it enables direct comparison with prior work.

\textbf{Baselines.}
We compare against published LiDAR completion methods (PVD, LODE,
LMSCNet, LiDiff, ScoreLiDAR, LiNeXt) that use standard SemanticKITTI GT
and output ${\sim}180{,}000$ points.
Our teacher outputs 20{,}000 points, and its v2 rows use the refined GT;
direct numerical comparison should be interpreted with caution.

\textbf{Hardware.}
All experiments run on a single NVIDIA RTX 4090 (24\,GB); the VAE
trains in 21\,h and the diffusion teacher for 30 epochs. End-to-end pipeline latency, covering encoder,
denoiser, an inference-time kNN densification (distinct from the offline
GT refinement), and device--host transfers, is $209$\,ms/frame
(4.78\,FPS; $129$\,ms encoder, $70$\,ms denoiser, $10$\,ms densification) at a 20{,}000-point input / 20{,}000-point target configuration.

\begin{figure*}[t]
\centering
\includegraphics[width=0.96\textwidth]{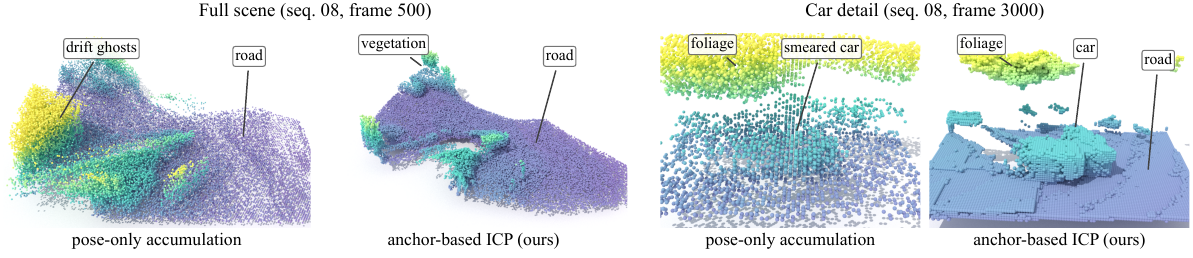}
\caption{Ground-truth refinement: pose-only accumulation vs.\ anchor-based
ICP (identical crops, cameras, and point budgets within each pair).
Drift-induced ghost surfaces and duplicated walls at scene scale (left
pair) are eliminated; in close-up (right pair), the car and road surface
become sharp after refinement.}
\label{fig:icp_scene}
\end{figure*}

\subsection{VAE Reconstruction}

Table~\ref{tbl:vae} summarises our VAE's performance.
The multi-token design achieves squared CD $0.120 \pm 0.026$\,m$^2$,
avoiding the codebook collapse (squared CD ${\sim}16$\,m$^2$) observed in
a single-codebook VQ-VAE baseline.
The 32-token setting balances reconstruction capacity against
decoder cost.

\begin{table}[ht]
\caption{VAE Performance on SemanticKITTI val (seq.~08, 50 frames).}
\label{tbl:vae}
\begin{center}
\begin{footnotesize}
\begin{tabular}{@{}lc@{}}
\toprule
\textbf{Metric} & \textbf{Value} \\
\midrule
Squared Chamfer Distance (CD) & $0.120 \pm 0.026$\,m$^2$ \\
Parameters & 7.1\,M \\
Inference time & 1.6\,ms/frame \\
Training time & 21\,h (100 epochs) \\
Decoded points & 8{,}000 \\
Latent tokens & 32 $\times$ 32-d (1{,}024-d total) \\
\bottomrule
\end{tabular}
\end{footnotesize}
\end{center}
\end{table}

\subsection{Ground Truth Refinement}
Table~\ref{tbl:gt} shows the effect of anchor-based ICP on the
teacher's measured squared CD.
The ${\sim}16\times$ squared-CD improvement is obtained with no hyperparameter
or architecture changes; as noted in Sec.~\ref{sec:discussion}, the refined
v2 references are also denser and more compact in extent, so much of the improvement
reflects the reference data rather than better predictions.

\begin{table}[ht]
\caption{Anchor-based ICP Ground Truth Refinement Results
(seq.~08, 4{,}071 frames $\times$ 20{,}000-point subsample, diffusion teacher).
Values are squared Chamfer distance (m$^2$), following the PVD / LiDiff
convention; GT variants are evaluation references (Sec.~\ref{sec:discussion}).}
\label{tbl:gt}
\begin{center}
\begin{footnotesize}
\begin{tabular}{@{}lcc@{}}
\toprule
\textbf{GT Variant} & \textbf{Squared CD (m$^2$) mean$\pm$std} & \textbf{Val Loss} \\
\midrule
v1 (original poses) & $0.396 \pm 0.090$ & 0.213 \\
v2 (anchor-based ICP) & $\mathbf{0.024 \pm 0.005}$ & 0.136 \\
\bottomrule
\end{tabular}
\end{footnotesize}
\end{center}
\end{table}

Beyond the ${\sim}16\times$ numerical gain, Fig.~\ref{fig:icp_scene} shows the visual signature of refinement:
duplicate surfaces disappear at scene scale, and object boundaries
sharpen markedly in close-up.

The structure of the ground truth refinement algorithm is illustrated in Fig.~\ref{fig:grtr_scheme}.

\subsection{Comparison with Baselines}
\label{sec:comparison}

We now place the diffusion teacher in context with published baselines
(Table~\ref{tbl:cd_compare}). Published rows are reported
as-is, evaluated on the full seq.~08 set (${\sim}4{,}071$ frames) under LiDiff's
protocol~\cite{LiDiff2024} (18{,}000-point input, 180{,}000-point output,
linear CD). With standard (v1) GT our teacher reaches linear CD $0.496$\,m,
comparable to LiDiff ($0.434$\,m). Our v2-GT squared CD ($0.024$\,m$^2$) uses a
tighter bbox and is \emph{not} unit-comparable to the published linear CDs; it
is shown for completeness.

\begin{table}[ht]
\caption{Chamfer Distance comparison on SemanticKITTI val. See
Sec.~\ref{sec:comparison} for protocol details. $^{\dagger}$Squared CD on a
tighter bounding box; not unit-comparable to the linear CDs above.}
\label{tbl:cd_compare}
\begin{center}
\begin{footnotesize}
\setlength{\tabcolsep}{4pt}
\begin{tabular}{@{}lcc@{}}
\toprule
\textbf{Method} & \textbf{Eval fr.} & \textbf{CD $\downarrow$} \\
\midrule
PVD~\cite{PVD2021} & ${\sim}4{,}071$ & 1.256\,m \\
LODE~\cite{LODE2023} & ${\sim}4{,}071$ & 1.029\,m \\
LMSCNet~\cite{LMSCNet2020} & ${\sim}4{,}071$ & 0.641\,m \\
LiDiff~\cite{LiDiff2024} & ${\sim}4{,}071$ & 0.434\,m \\
ScoreLiDAR~\cite{ScoreLiDAR2025} & ${\sim}4{,}071$ & 0.406\,m \\
LiNeXt~\cite{LiNeXt2025} & ${\sim}4{,}071$ & 0.149\,m \\
\midrule
Ours, v1 GT (linear) & 4{,}071 & $0.496$\,m \\
Ours, v1 GT (squared) & 4{,}071 & $0.396 \pm 0.090$\,m$^2$ \\
Ours, v2 GT (squared) & 4{,}071 & $0.024 \pm 0.005$\,m$^2$\,$^{\dagger}$ \\
\bottomrule
\end{tabular}
\end{footnotesize}
\end{center}
\end{table}

\textbf{Inference latency.}
Our single-step end-to-end pipeline runs at $209$\,ms/frame on an
RTX~4090. Timed on the same hardware from native $360^\circ$ scans, LiDiff's 50-step
DDPM sampler takes $20.56$\,s/frame ($98\times$ slower) and ScoreLiDAR's
released 8-step implementation $4.72$\,s/frame ($23\times$ slower),
consistent with the $30.55$\,s and $5.16$\,s their authors report on their
own hardware. Only the non-diffusion LiNeXt~\cite{LiNeXt2025} is comparable
($167$\,ms/frame); ours is, to our knowledge, the fastest published
diffusion-based completion method.

\section{Discussion}
\label{sec:discussion}

\textbf{Data quality dominates architecture.}
The ${\sim}16\times$ squared-CD improvement from GT refinement (v1$\to$v2)
with zero model changes is the single largest gain in our study; at
this denoiser scale supervision quality, not capacity, is the binding
constraint.
Note that v2 GT is also considerably denser and spans a tighter forward-facing
bounding box, so part of the CD reduction reflects denser reference points
and smaller spatial extent rather than better predictions. A $2{\times}2$
cross-evaluation supports this reading: the v1-trained teacher scores
$0.362$/$0.026$\,m$^2$ against the v1/v2 references versus $0.396$/$0.024$
for the v2-trained teacher, attributing most of the ratio to the reference
change, with a consistent $8.1\%$ paired per-frame gain from matched supervision at a fixed
reference (paired 95\% CI $[8.0, 8.2]\%$, $n{=}4{,}071$).

\textbf{Scaffold-free deployment.} With scaffolds built only from the
partial scan (no GT access), the teacher is out-of-distribution, but a
45-minute mixed-scaffold fine-tune reaches a squared CD of $0.97$\,m$^2$
under this harder protocol, lower than re-evaluated LiDiff
($2.32$\,m$^2$) and ScoreLiDAR ($2.10$\,m$^2$), both re-run from native
scans on the same frames, cropped to the same volume, and scored as
squared CD against v2 GT; a companion manuscript studies this in depth.

\textbf{Multi-token design avoids collapse.}
A single-codebook VQ-VAE~\cite{VQVAE2017} baseline plateaued at squared
CD ${\sim}16$\,m$^2$ (flat validation loss from the first epoch) due to
codebook collapse; the multi-token VAE ($0.120$\,m$^2$) avoids it by
distributing information across 32 specialised tokens.

\textbf{Limitations.}
Evaluation is confined to SemanticKITTI; indoor and off-road
generalisation is untested.
The VAE remains a standalone reconstruction module; latent-bottleneck
integration is future work.
Full DDPM reverse sampling was unstable in our setting; single-step $x_0$
inference limits output diversity.


\section{Conclusion}

We presented CloudDiffusion: its multi-token Gaussian VAE compresses
scene-scale LiDAR point clouds to squared CD $0.120 \pm 0.026$\,m$^2$
at 7.1\,M parameters, and anchor-based ICP ground truth refinement
improves the diffusion teacher's measured squared Chamfer distance by
${\sim}16\times$
($0.396 \to 0.024$\,m$^2$) without architectural change, most of it
attributable to the refined evaluation substrate (Sec.~\ref{sec:discussion}).
Our single-step $x_0$ teacher, operating directly in coordinate space,
completes scenes at $23\text{--}98\times$ lower inference latency than
iterative diffusion baselines. Future work will integrate the VAE as a
latent bottleneck for the denoiser, pursue scaffold-free single-step
completion, and extend evaluation beyond urban driving.

\section*{Acknowledgements} 
Research reported in this publication was financially supported by the RSF grant No. 24-41-02039.



\begin{thebibliography}{10}
\providecommand{\url}[1]{#1}
\csname url@samestyle\endcsname
\providecommand{\newblock}{\relax}
\providecommand{\bibinfo}[2]{#2}
\providecommand{\BIBentrySTDinterwordspacing}{\spaceskip=0pt\relax}
\providecommand{\BIBentryALTinterwordstretchfactor}{4}
\providecommand{\BIBentryALTinterwordspacing}{\spaceskip=\fontdimen2\font plus
\BIBentryALTinterwordstretchfactor\fontdimen3\font minus \fontdimen4\font\relax}
\providecommand{\BIBforeignlanguage}[2]{{%
\expandafter\ifx\csname l@#1\endcsname\relax
\typeout{** WARNING: IEEEtran.bst: No hyphenation pattern has been}%
\typeout{** loaded for the language `#1'. Using the pattern for}%
\typeout{** the default language instead.}%
\else
\language=\csname l@#1\endcsname
\fi
#2}}
\providecommand{\BIBdecl}{\relax}
\BIBdecl

\bibitem{lederman2009haptic}
S.~J. Lederman and R.~L. Klatzky, ``Haptic perception: A tutorial,'' \emph{Attention, Perception, \& Psychophysics}, vol.~71, no.~7, pp. 1439--1459.

\bibitem{7780460}
J.~Redmon, S.~Divvala, R.~Girshick, and A.~Farhadi, ``{ You Only Look Once: Unified, Real-Time Object Detection },'' in \emph{Proc. IEEE Conf. on Computer Vision and Pattern Recognition (CVPR)}, 2016, pp. 779--788.

\bibitem{yu2205coca}
J.~Yu, Z.~Wang, V.~Vasudevan, L.~Yeung, M.~Seyedhosseini, and Y.~Wu, ``Coca: Contrastive captioners are image-text foundation models,'' 2022, arXiv:2205.01917.

\bibitem{drehwald2023one}
M.~S. Drehwald, S.~Eppel, J.~Li, H.~Hao, and A.~Aspuru-Guzik, ``One-shot recognition of any material anywhere using contrastive learning with physics-based rendering,'' in \emph{Proc. of the IEEE/CVF Int. Conf. on Computer Vision}, 2023, pp. 23\,524--23\,533.

\bibitem{bensmaia2003vibrations}
S.~J. Bensma{\"\i}a and M.~Hollins, ``The vibrations of texture,'' \emph{Somatosensory \& motor research}, vol.~20, no.~1, pp. 33--43.

\bibitem{pacchierotti2017wearable}
C.~Pacchierotti, S.~Sinclair, M.~Solazzi, A.~Frisoli, V.~Hayward, and D.~Prattichizzo, ``Wearable haptic systems for the fingertip and the hand: taxonomy, review, and perspectives,'' \emph{IEEE transactions on haptics}, vol.~10, no.~4, pp. 580--600, 2017.

\bibitem{kim2020thermal}
S.-W. Kim, S.~H. Kim, C.~S. Kim, K.~Yi, J.-S. Kim, B.~J. Cho, and Y.~Cha, ``Thermal display glove for interacting with virtual reality,'' \emph{Scientific reports}, vol.~10, no.~1, p. 11403, 2020.

\bibitem{salisbury2004haptic}
K.~Salisbury, F.~Conti, and F.~Barbagli, ``Haptic rendering: introductory concepts,'' \emph{IEEE computer graphics and applications}, vol.~24, no.~2, pp. 24--32, 2004.

\bibitem{altamirano2020}
M.~Altamirano~Cabrera, J.~Heredia, and D.~Tsetserukou, ``Tactile perception of objects by the user's palm for the development of multi-contact wearable tactile displays,'' in \emph{Proc. Int. Conf. EuroHaptics 2020}, 2020, pp. 51--59.

\bibitem{culbertson2013generating}
H.~Culbertson, J.~Unwin, B.~E. Goodman, and K.~J. Kuchenbecker, ``Generating haptic texture models from unconstrained tool-surface interactions,'' in \emph{Proc. World Haptics Conference (WHC)}, 2013, pp. 295--300.

\bibitem{sinapov2011interactive}
J.~Sinapov, T.~Bergquist, C.~Schenck, U.~Ohiri, S.~Griffith, and A.~Stoytchev, ``Interactive object recognition using proprioceptive and auditory feedback,'' \emph{The International Journal of Robotics Research}, vol.~30, no.~10, pp. 1250--1262, 2011.

\bibitem{kuchenbecker2006improving}
K.~J. Kuchenbecker, J.~Fiene, and G.~Niemeyer, ``Improving contact realism through event-based haptic feedback,'' \emph{IEEE transactions on visualization and computer graphics}, vol.~12, no.~2, pp. 219--230.

\bibitem{cabrera2024}
M.~A. Cabrera, M.~H. Khan, A.~Alabbas, L.~Moreno, I.~Tokmurziyev, and D.~Tsetserukou, ``Musinger: Communication of music over a distance with wearable haptic display and touch sensitive surface,'' 2024, arXiv:2410.16202.

\bibitem{gao2016deep}
Y.~Gao, L.~A. Hendricks, K.~J. Kuchenbecker, and T.~Darrell, ``Deep learning for tactile understanding from visual and haptic data,'' in \emph{Proc. IEEE Int. Conf. on robotics and automation (ICRA)}.\hskip 1em plus 0.5em minus 0.4em\relax IEEE, 2016, pp. 536--543.

\bibitem{yuan2017gelsight}
W.~Yuan, S.~Dong, and E.~H. Adelson, ``Gelsight: High-resolution robot tactile sensors for estimating geometry and force,'' \emph{Sensors}, vol.~17, no.~12, p. 2762, 2017.

\bibitem{owens2016visually}
A.~Owens, P.~Isola, J.~McDermott, A.~Torralba, E.~H. Adelson, and W.~T. Freeman, ``Visually indicated sounds,'' in \emph{Proc. IEEE Conf. on computer vision and pattern recognition}, 2016, pp. 2405--2413.

\bibitem{peiris2017thermovr}
R.~L. Peiris, W.~Peng, Z.~Chen, L.~Chan, and K.~Minamizawa, ``Thermovr: Exploring integrated thermal haptic feedback with head mounted displays,'' in \emph{Proc. Conf. on Human Factors in Computing Systems (CHI)}, 2017, pp. 5452--5456.

\bibitem{erickson2020multimodal}
Z.~Erickson, E.~Xing, B.~Srirangam, S.~Chernova, and C.~C. Kemp, ``Multimodal material classification for robots using spectroscopy and high resolution texture imaging,'' in \emph{Proc. IEEE/RSJ Int. Conf. on Intelligent Robots and Systems (IROS)}.\hskip 1em plus 0.5em minus 0.4em\relax IEEE, 2020, pp. 10\,452--10\,459.

\bibitem{li2023blip}
J.~Li, D.~Li, S.~Savarese, and S.~Hoi, ``Blip-2: bootstrapping language-image pre-training with frozen image encoders and large language models,'' in \emph{Proc. 40th Int. Conf. on Machine Learning (ICML)}.\hskip 1em plus 0.5em minus 0.4em\relax JMLR.org, 2023.

\bibitem{alayrac2022flamingo}
J.-B. Alayrac, J.~Donahue, P.~Luc, A.~Miech, I.~Barr, Y.~Hasson, K.~Lenc, A.~Mensch, K.~Millicah, M.~Reynolds \emph{et~al.}, ``Flamingo: a visual language model for few-shot learning,'' in \emph{Proc. 36th Int. Conf. on Neural Information Processing Systems (NIPS)}, 2022.

\bibitem{peng2023kosmos}
Z.~Peng, W.~Wang, L.~Dong, Y.~Hao, S.~Huang, S.~Ma, and F.~Wei, ``Kosmos-2: Grounding multimodal large language models to the world,'' 2023, arXiv:2306.14824.

\bibitem{deitke2024molmopixmoopenweights}
M.~Deitke, C.~Clark, S.~Lee, R.~Tripathi, Y.~Yang, J.~S. Park, M.~Salehi, N.~Muennighoff, K.~Lo, L.~Soldaini,  \emph{et~al.}, ``Molmo and pixmo: Open weights and open data for state-of-the-art vision-language models,'' 2024, arXiv:2409.17146.

\bibitem{wang2024qwen2}
P.~Wang, S.~Bai, S.~Tan, S.~Wang, Z.~Fan, J.~Bai, K.~Chen, X.~Liu, J.~Wang, W.~Ge, Y.~Fan, K.~Dang, M.~Du, X.~Ren, R.~Men, D.~Liu, C.~Zhou, J.~Zhou, and J.~Lin, ``Qwen2-vl: Enhancing vision-language model's perception of the world at any resolution,'' 2024, arXiv:2409.12191.

\bibitem{a10010015}
O.~C. Santos, ``Toward personalized vibrotactile support when learning motor skills,'' \emph{Algorithms}, vol.~10, no.~1, 2017.

\bibitem{kristjansson2016designing}
{\'A}.~Kristj{\'a}nsson, A.~Moldoveanu, {\'O}.~I. J{\'o}hannesson, O.~Balan, S.~Spagnol, V.~V. Valgeirsd{\'o}ttir, and R.~Unnthorsson, ``Designing sensory-substitution devices: Principles, pitfalls and potential 1,'' \emph{Restorative neurology and neuroscience}, vol.~34, no.~5, pp. 769--787, 2016.

\bibitem{rasouli2018extreme}
M.~Rasouli, Y.~Chen, A.~Basu, S.~L. Kukreja, and N.~V. Thakor, ``An extreme learning machine-based neuromorphic tactile sensing system for texture recognition,'' \emph{IEEE transactions on biomedical circuits and systems}, vol.~12, no.~2, pp. 313--325, 2018.

\bibitem{shi2020evaluation}
W.~Shi, M.~Li, J.~Guo, and K.~Zhai, ``Evaluation of road service performance based on human perception of vibration while driving vehicle,'' \emph{Journal of Advanced Transportation}, vol. 2020, no.~1, p. 8825355, 2020.

\end{thebibliography}


\begin{thebibliography}{99}
\bibitem{EmbodiedDiffusion2026}
I.~Zhura \textit{et al.},
``EmbodiedDiffusion: End-to-end traversability-guided visual diffusion for heterogeneous robot navigation,''
2025, arXiv:2512.02851.
\bibitem{DiffusionAnything2026}
I.~Zhura \textit{et al.},
``DiffusionAnything: End-to-end in-context diffusion learning for unified navigation and pre-grasp motion,''
2026, arXiv:2603.26322.
\bibitem{DiffusionRL2026}
M.~Makarova, Q.~Liu, and D.~Tsetserukou,
``DiffusionRL: Efficient training of diffusion policies for robotic grasping using RL-adapted large-scale datasets,''
2025, arXiv:2505.18876.
\bibitem{ImpedanceDiffusion2026}
F.~Batool, Y.~Yaqoot, M.~A.~Mustafa, R.~A.~Khan, A.~Fedoseev, and D.~Tsetserukou,
``ImpedanceDiffusion: Diffusion-based global path planning for UAV swarm navigation with generative impedance control,''
in \textit{Proc. IEEE Int. Conf. Unmanned Aircraft Systems (ICUAS)}, 2026, pp.~537--544.
\bibitem{LMSCNet2020}
L.~Rold\~{a}o, R.~de Charette, and A.~Verroust-Blondet,
``LMSCNet: Lightweight multiscale 3D semantic completion,''
in \textit{Proc. Int. Conf. 3D Vision (3DV)}, 2020, pp.~111--119.
\bibitem{VoxFormer2023}
Y.~Li \textit{et al.},
``VoxFormer: Sparse voxel transformer for camera-based 3D semantic scene completion,''
in \textit{Proc. IEEE/CVF Conf. Computer Vision and Pattern Recognition (CVPR)}, 2023, pp.~9087--9098.
\bibitem{CGFormer2024}
Z.~Yu \textit{et al.},
``Context and geometry aware voxel transformer for semantic scene completion,''
in \textit{Proc. Advances in Neural Information Processing Systems (NeurIPS)}, 2024.
\bibitem{SGN2024}
J.~Mei \textit{et al.},
``Camera-based 3D semantic scene completion with sparse guidance network,''
\textit{IEEE Trans. Image Process.}, vol.~33, pp.~5468--5481, 2024.
\bibitem{DepthSSC2025}
J.~Yao, J.~Zhang, X.~Pan, T.~Wu, and C.~Xiao,
``DepthSSC: Monocular 3D semantic scene completion via depth-spatial alignment and voxel adaptation,''
in \textit{Proc. IEEE/CVF Winter Conf. Applications of Computer Vision (WACV)}, 2025, pp.~2154--2163.
\bibitem{LODE2023}
P.~Li \textit{et al.},
``LODE: Locally conditioned eikonal implicit scene completion from sparse LiDAR,''
in \textit{Proc. IEEE Int. Conf. Robotics and Automation (ICRA)}, 2023, pp.~8269--8276.
\bibitem{LiNeXt2025}
W.~He \textit{et al.},
``LiNeXt: Revisiting LiDAR completion with efficient non-diffusion architectures,''
in \textit{Proc. AAAI Conf. Artificial Intelligence (AAAI)}, 2026, pp.~4672--4680.
\bibitem{R2DM2024}
K.~Nakashima and R.~Kurazume,
``LiDAR data synthesis with denoising diffusion probabilistic models,''
in \textit{Proc. IEEE Int. Conf. Robotics and Automation (ICRA)}, 2024, pp.~14724--14731.
\bibitem{LiDMs2024}
H.~Ran, V.~Guizilini, and Y.~Wang,
``Towards realistic scene generation with LiDAR diffusion models,''
in \textit{Proc. IEEE/CVF Conf. Computer Vision and Pattern Recognition (CVPR)}, 2024, pp.~14738--14748.
\bibitem{DiffComplete2024}
R.~Chu \textit{et al.},
``DiffComplete: Diffusion-based generative 3D shape completion,''
in \textit{Proc. Advances in Neural Information Processing Systems (NeurIPS)}, 2023.
\bibitem{LiDiff2024}
L.~Nunes, R.~Marcuzzi, B.~Mersch, J.~Behley, and C.~Stachniss,
``Scaling diffusion models to real-world 3D LiDAR scene completion,''
in \textit{Proc. IEEE/CVF Conf. Computer Vision and Pattern Recognition (CVPR)}, 2024, pp.~14770--14780.
\bibitem{ScoreLiDAR2025}
S.~Zhang \textit{et al.},
``Distilling diffusion models to efficient 3D LiDAR scene completion,''
in \textit{Proc. IEEE/CVF Int. Conf. Computer Vision (ICCV)}, 2025, pp.~5007--5016.
\bibitem{DistDPO2025}
A.~Zhao \textit{et al.},
``Diffusion distillation with direct preference optimization for efficient
3D LiDAR scene completion,'' 2025, arXiv:2504.11447.
\bibitem{LiDPM2025}
T.~Martyniuk, G.~Puy, A.~Boulch, R.~Marlet, and R.~de~Charette,
``LiDPM: Rethinking point diffusion for LiDAR scene completion,''
in \textit{Proc. IEEE Intelligent Vehicles Symposium (IV)}, 2025, pp.~555--560.
\bibitem{PointNet2017}
C.~R.~Qi, H.~Su, K.~Mo, and L.~J.~Guibas,
``PointNet: Deep learning on point sets for 3D classification and segmentation,''
in \textit{Proc. IEEE Conf. Computer Vision and Pattern Recognition (CVPR)}, 2017, pp.~652--660.
\bibitem{VQVAE2017}
A.~van~den~Oord, O.~Vinyals, and K.~Kavukcuoglu,
``Neural discrete representation learning,''
in \textit{Proc. Advances in Neural Information Processing Systems (NeurIPS)}, 2017, pp.~6306--6315.
\bibitem{Perceiver2021}
A.~Jaegle, F.~Gimeno, A.~Brock, O.~Vinyals, A.~Zisserman, and J.~Carreira,
``Perceiver: General perception with iterative attention,''
in \textit{Proc. Int. Conf. Machine Learning (ICML)}, 2021, pp.~4651--4664.
\bibitem{Shape2VecSet2023}
B.~Zhang, J.~Tang, M.~Nie{\ss}ner, and P.~Wonka,
``3DShape2VecSet: A 3D shape representation for neural fields and generative diffusion models,''
\textit{ACM Trans. Graph.}, vol.~42, no.~4, pp.~92:1--92:16, 2023.
\bibitem{nuCraft2024}
B.~Zhu, Z.~Wang, and H.~Li,
``nuCraft: Crafting high resolution 3D semantic occupancy for unified 3D scene understanding,''
in \textit{Proc. Eur. Conf. Computer Vision (ECCV)}, 2024, pp.~125--141.
\bibitem{PTv3}
X.~Wu \textit{et al.},
``Point Transformer V3: Simpler, faster, stronger,''
in \textit{Proc. IEEE/CVF Conf. Computer Vision and Pattern Recognition (CVPR)}, 2024, pp.~4840--4851.
\bibitem{Sonata}
X.~Wu \textit{et al.},
``Sonata: Self-supervised learning of reliable point representations,''
in \textit{Proc. IEEE/CVF Conf. Computer Vision and Pattern Recognition (CVPR)}, 2025, pp.~22193--22204.
\bibitem{DDPM2020}
J.~Ho, A.~Jain, and P.~Abbeel,
``Denoising diffusion probabilistic models,''
in \textit{Proc. Advances in Neural Information Processing Systems (NeurIPS)}, vol.~33, 2020, pp.~6840--6851.
\bibitem{iDDPM2021}
A.~Nichol and P.~Dhariwal,
``Improved denoising diffusion probabilistic models,''
in \textit{Proc. Int. Conf. Machine Learning (ICML)}, 2021, pp.~8162--8171.
\bibitem{SemanticKITTI2019}
J.~Behley \textit{et al.},
``SemanticKITTI: A dataset for semantic scene understanding of LiDAR sequences,''
in \textit{Proc. IEEE/CVF Int. Conf. Computer Vision (ICCV)}, 2019, pp.~9297--9307.
\bibitem{PVD2021}
L.~Zhou, Y.~Du, and J.~Wu,
``3D shape generation and completion through point-voxel diffusion,''
in \textit{Proc. IEEE/CVF Int. Conf. Computer Vision (ICCV)}, 2021, pp.~5826--5835.
\end{thebibliography}
\end{document}